\documentclass[runningheads]{llncs}
\usepackage{bbding}
\usepackage{graphicx}
\usepackage{booktabs}
\usepackage{hyperref}
\usepackage{multirow}
\usepackage{amsmath}
\usepackage{amssymb}
\usepackage{mathtools}
\usepackage{colortbl}
\usepackage{xcolor}
\usepackage{color}
\usepackage{caption,subcaption}
\definecolor{lm_purple_low}{RGB}{240,240,248}
\definecolor{lm_purple}{RGB}{227,227,240}
\definecolor{lm_purple_mid}{RGB}{233,222,254}
\definecolor{lm_red}{RGB}{230,36,43}
\definecolor{cblue}{rgb}{0.21,0.49,0.74}
\usepackage{todonotes}
\usepackage[marginal]{footmisc}

\usepackage{xspace}

\begin{document}
\title{Uncertainty-Aware Multi-Expert Knowledge Distillation for Imbalanced Disease Grading}
%
\titlerunning{UMKD}
%
\author{Shuo Tong\inst{1,4,5,*}
\and Shangde Gao\inst{2,3,4,5,*}
\and Ke Liu\inst{2}
\and Zihang Huang\inst{6}
\and Hongxia Xu\inst{3}
\and Haochao Ying\inst{1,4,}\Envelope
\and Jian Wu \inst{1,3,4,5}}

\authorrunning{Shangde Gao, et al.}

\institute{School of Public Health, Zhejiang University
\and College of Computer Science and Technology, Zhejiang University
\and Liangzhu Laboratory and WeDoctor Cloud
\and 
State Key Laboratory of Transvascular Implantation Devices, The Second Affiliated Hospital Zhejiang University School of Medicine
\and
Zhejiang Key Laboratory of Medical Imaging Artificial Intelligence
\and
School of Electronic Information and Communications, Huazhong University of Science and Technology
\\
\email{\{tongshuo, gaosde, lk2017, Einstein, haochaoying, wujian2000\}@zju.edu.cn}\\
}
\maketitle              
\footnotetext{*: Equal-contribution authors.\\
\Envelope: Corresponding author.}
\begin{abstract}
Automatic disease image grading is a significant application of artificial intelligence for healthcare, enabling faster and more accurate patient assessments. 
However, domain shifts, which are exacerbated by data imbalance, introduce bias into the model, posing deployment difficulties in clinical applications. 
To address the problem, we propose a novel \textbf{U}ncertainty-aware \textbf{M}ulti-experts \textbf{K}nowledge \textbf{D}istillation (UMKD) framework to transfer knowledge from multiple expert models to a single student model. 
Specifically, to extract discriminative features, UMKD decouples task-agnostic and task-specific features with shallow and compact feature alignment in the feature space. 
At the output space, an uncertainty-aware decoupled distillation (UDD) mechanism dynamically adjusts knowledge transfer weights based on expert model uncertainties, ensuring robust and reliable distillation.
Additionally, UMKD also tackles the problems of model architecture heterogeneity and distribution discrepancies between source and target domains, which are inadequately tackled by previous KD approaches.
Extensive experiments on histology prostate grading (\textit{SICAPv2}) and fundus image grading (\textit{APTOS}) demonstrate that UMKD achieves a new state-of-the-art in both source-imbalanced and target-imbalanced scenarios, offering a robust and practical solution for real-world disease image grading.
\end{abstract}
\section{Introduction}

Image-driven disease grading systems are pivotal for enhancing clinical decision-making efficiency~\cite{litjens2017survey,xie2021survey}, especially for diabetic retinopathy (DR) and prostate cancer.
Early-stage precise grading significantly improves patient prognosis (\textit{e.g.}, timely intervention reduces blindness risk by 90\% in DR patients)~\cite{porwal2020idrid,bulten2022artificial}.
However, traditional grading is largely limited by challenges such as differences in subjective expert judgment and difficulty in identifying subtle pathologic features. 
Clinical evidence reports a 40\% inter-observer variability in Gleason scoring and a misdiagnosis rate of more than 25\% for early DR microaneurysms.

Recently, medical efficiency has been enhanced by the rapid development of artificial intelligence-based automatic disease grading systems~\cite{cheng2023robust,gao2023contrastive}. For DR grading, 
Dai \textit{et al.}~\cite{dai2021deep} developed DeepDR to detect early-to-late stages of diabetic retinopathy.
Wang \textit{et al.}~\cite{wang2023ord2seq} reformulated DR grading as sequence prediction, resolving ambiguous boundary issues effectively.
On the other hand, for prostate cancer grading, Morpho-Grader~\cite{claudio2024mapping} disentangles glandular morphology from stromal textures.
BayeSeg~\cite{gao2023bayeseg} employs variational inference to separate structural invariants from texture variations, significantly improving model robustness.

\begin{figure}[t!]
    \centering
    \begin{subfigure}[b]{0.24\textwidth}
        \includegraphics[width=\textwidth]{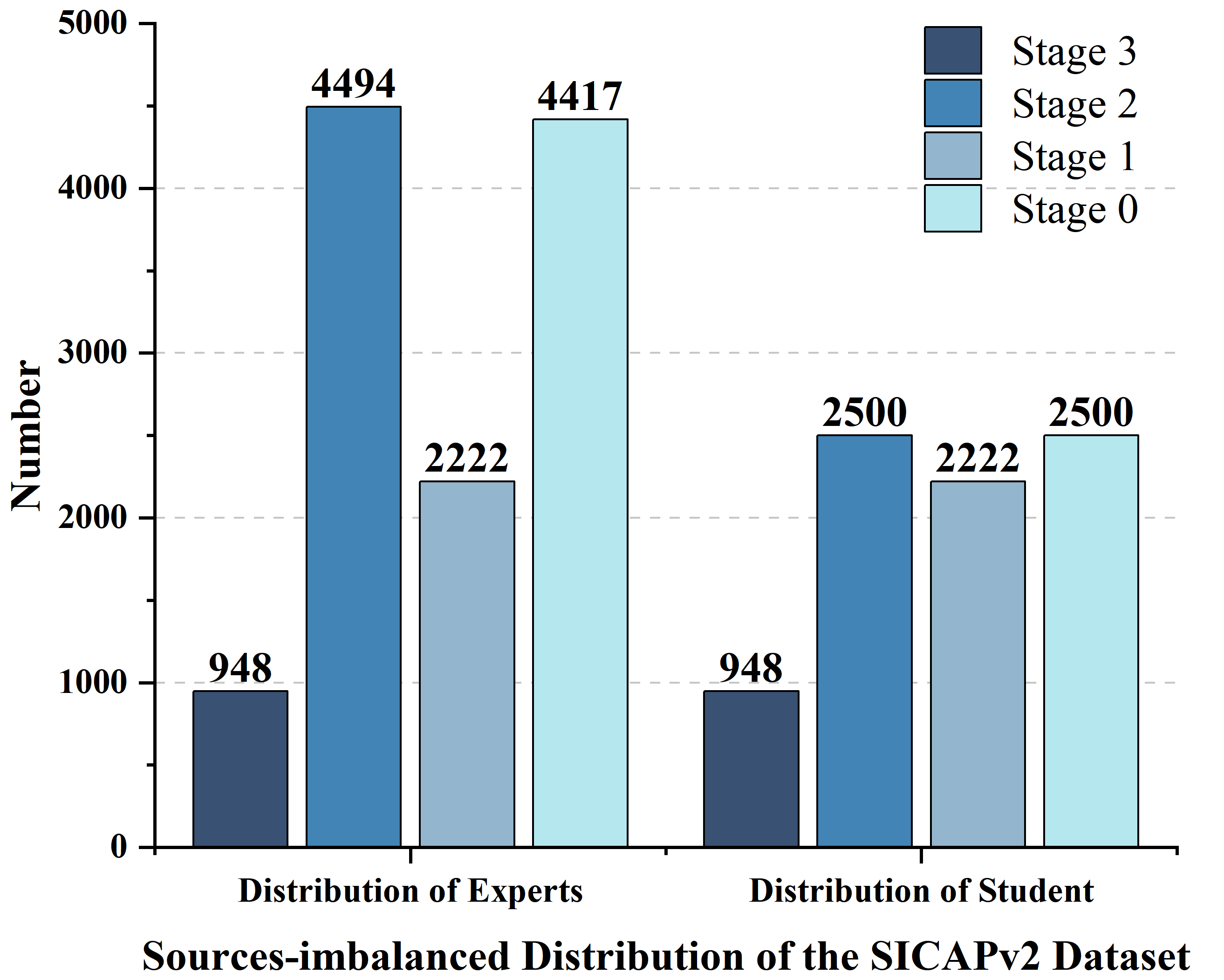 }
        \label{subfig:image1}
    \end{subfigure}
    \hfill
    \begin{subfigure}[b]{0.24\textwidth}
        \includegraphics[width=\textwidth]{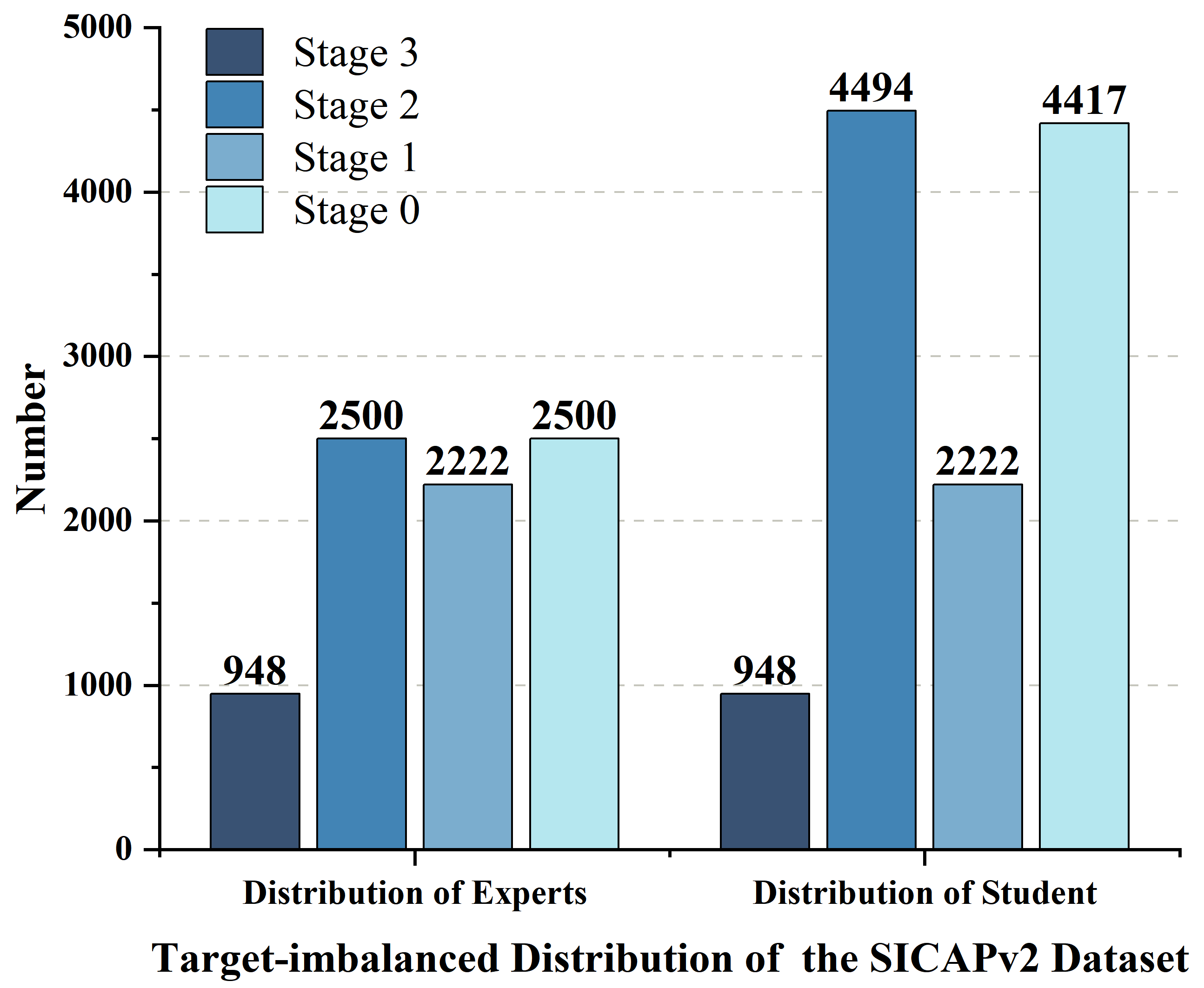}
        \label{subfig:image2}
    \end{subfigure}
    \hfill
    \begin{subfigure}[b]{0.24\textwidth}
        \includegraphics[width=\textwidth]{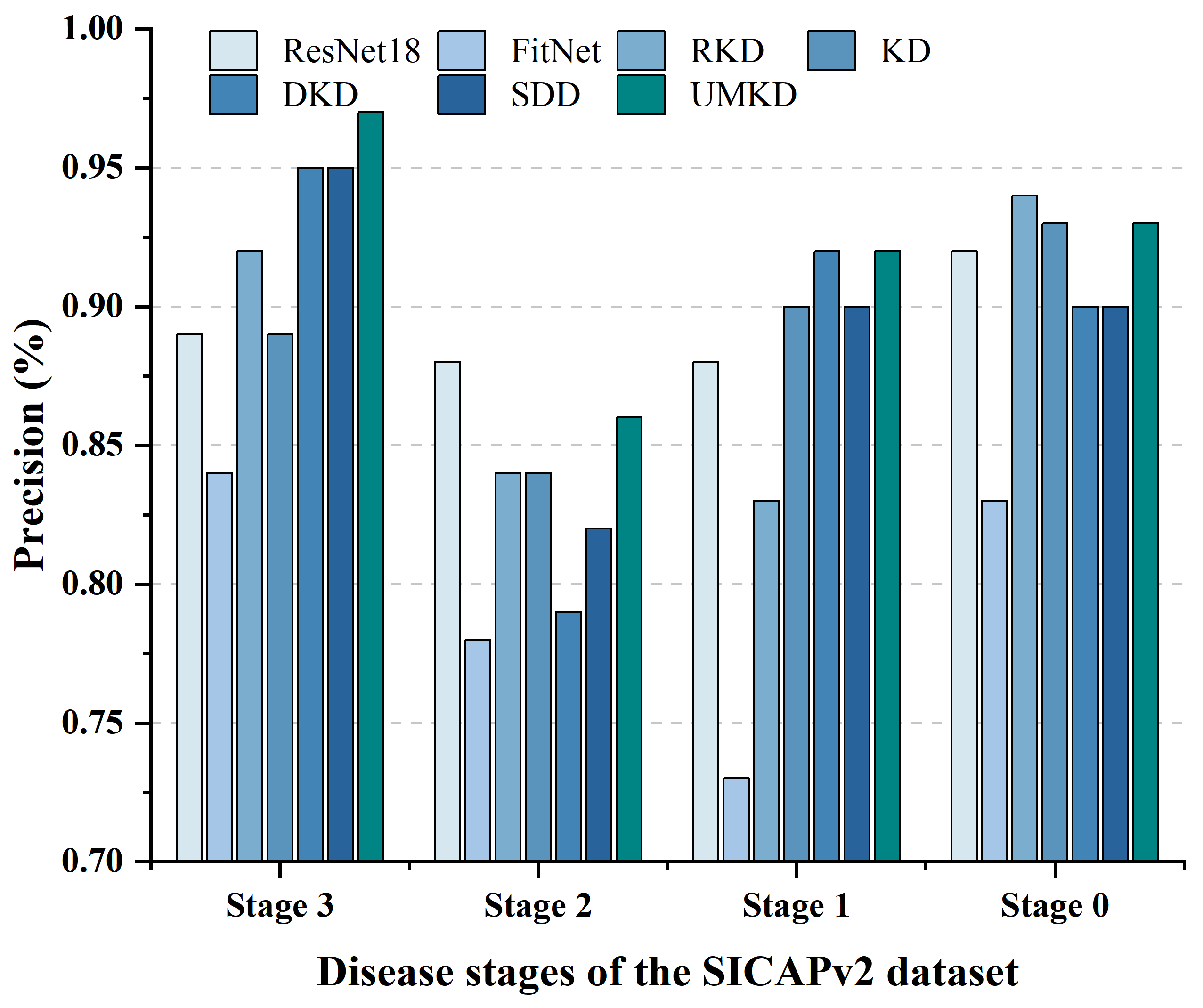}
        \label{subfig:image3}
    \end{subfigure}
    \hfill
    \begin{subfigure}[b]{0.24\textwidth}
        \includegraphics[width=\textwidth]{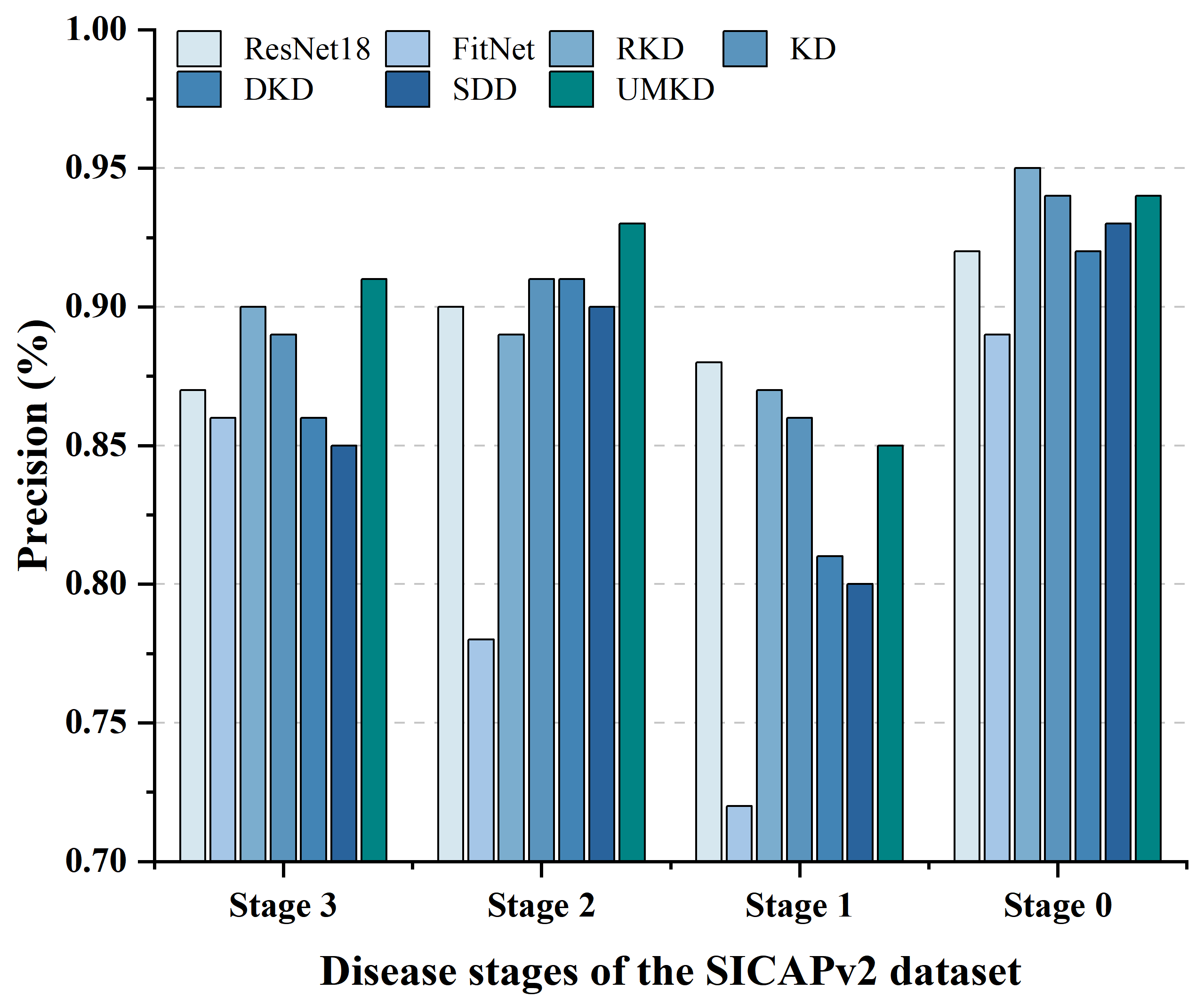}
        \label{subfig:image4}
    \end{subfigure}
    \caption{Domain shifts between source and target data (left) and the performance of methods (right) for \textit{sources-imbalanced} and \textit{target-imbalanced} KD tasks.}
    \label{fig:visualization}
\end{figure}

Despite progress in disease grading methods~\cite{mohan2023drfl,bulten2022artificial}, their deployment in the clinical applications is still limited by domain shifts.
Especially, imbalanced data exacerbate domain shifts (differences between source and target domain distributions).
As demonstrated in the SICAPv2 dataset (Fig.\ref{fig:visualization} left), stage III prostate cancer samples constitute 8\% of the cohort, leading models overfit to the majority class.
This imbalance is equally pronounced in DR grading, where scarce early-stage lesions reduce microaneurysm detection sensitivity by over 30\%~\cite{aptos2019-blindness-detection}.
Multi-expert knowledge distillation (MKD)~\cite{gao2024collaborative,hao2024one,gao2024ka}, a technique that improves the generalization of the student for minority class samples by transferring expert model knowledge. Due to its robustness to domain shifts, MKD has been applied to address class imbalance in natural images~\cite{wei2024scaled,li2024enhancing}, but its study in disease image grading remains underexplored.

In this paper, we propose a novel \textbf{U}ncertainty-aware \textbf{M}ulti-experts \textbf{K}nowledge \textbf{D}istillation (UMKD) framework to tackle the problem of class imbalance.
Specifically, to decouple the structural and semantic information of image representation, we design two feature alignment mechanisms: shallow feature alignment (SFA) and compact feature alignment (CFA).
SFA generalizes the alignment between expert and student features by multi-scale low-pass filtering, thereby preserving structural information (task-agnostic features) of disease images. 
CFA maps the features of the expert and student models to a common spherical space, allowing the student model to learn grading-related feature knowledge from each expert.
We design an uncertainty-aware decoupled distillation (UDD) mechanism at the output space, automatically detecting uncertainties in the expert model caused by class imbalance.
Via uncertainty metrics, the student model dynamically adjusts knowledge transfer weights, reducing bias propagation and ensuring a more robust and reliable knowledge transfer process.
Experimental results show that our method significantly outperforms existing multi-expert distillation approaches in fundus and prostate disease image grading tasks. Especially for imbalanced class and heterogeneous models, UMKD achieves much more reliable knowledge transfer as shown in Fig.~\ref{fig:visualization} right.
\section{Method}
\begin{figure*}[t!]
    \centering
    \includegraphics[width=0.98\linewidth]{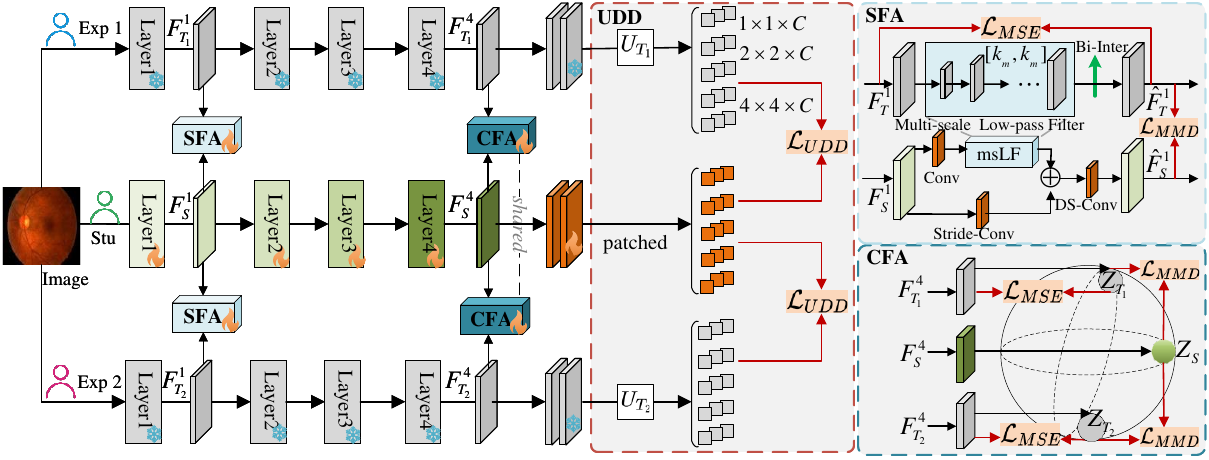}
    \caption{Model of uncertainty-aware multi-expert knowledge distillation. 
    }
    \label{fig:umkd}
\end{figure*}

\textbf{Model of the uncertainty-aware multi-expert knowledge distillation.}
We aim to distill the knowledge from multi-expert models into a target student model for imbalanced disease grading tasks. 
As shown in Fig.~\ref{fig:umkd}, our framework includes shallow feature alignment (SFA), compact feature alignment (CFA), and uncertainty-aware decoupled distillation (UDD). First, we will introduce the feature alignment loss, which will be reused by the SFA and CFA modules. Next, a detailed description of each module will be provided.

\textbf{Maximum Mean Discrepancy and Reconstruction Loss.}
\label{sec3:mmd_mse}
To measure the distribution differences between the student's features and those of each expert for feature alignment in the feature space, we employ Maximum Mean Discrepancy (MMD)~\cite{gao2024collaborative}.
The MMD distance is calculated as follows:
\begin{equation}
\label{eq:mmd}
\small
\mathcal{L}_\text{MMD}= \frac{1}{B}\sum_{t=1}^{N}\left\| \sum_{i=1}^{B} \phi\left(\hat{F}_{T_t}^{i}\right)- \sum_{j=1}^{B} \phi \left(\hat{F}_{S}^{j}\right)\right\|_{2}^{2},
\end{equation}
where $\phi$ is an explicit mapping function, $\hat{F}_{T_t}$ and $\hat{F}_{S}$ represent the features of expert $T_t$ and student $S$ after projection, respectively, $B$ is the batch size, and $N$ is the number of experts.
Meanwhile, to ensure that the expert models remain unchanged due to privacy constraints, the reconstruction loss \(\mathcal{L}_\text{MSE}\) is used to measure the changes in the expert models before and after feature alignment:
\begin{equation}
\label{eq:mse}
\small
\mathcal{L}_\text{MSE} = \sum_{t=1}^{N} \left\| F_{T_t} - \hat{F}_{T_t} \right\|_2^2,
\end{equation}
where \(F_{T_t}\) represents the original features of the expert model \(T_t\) before alignment, and \(\hat{F}_{T_t}\) denotes the decoded features of the expert model \(T_t\) after alignment.
By aggregating the alignment loss and the reconstruction loss, the total loss for feature alignment can be expressed as:
\begin{equation}
\label{total_loss}
\mathcal{L}_\text{FA} = \mathcal{L}_\text{MMD} + \mathcal{L}_\text{MSE}.
\end{equation}
\textbf{Shallow Feature Alignment.}
\label{sec3:sfa}
SFA preserves the main structural information of given images while removing noise and high-frequency details, ensuring consistency in generalized feature learning between expert models and student models~\cite{hao2024one}. Specifically, given shallow-layer feature representations between heterogeneous expert-student models, we propose to align features with multi-scale low-pass filtering from a frequency domain. 
For each expert feature \(F_{T_t}\), we adopt traditional average pooling as the low-pass filter and construct multi-scale filters by adjusting different kernel sizes and strides across multiple groups to accommodate different cutoff frequencies. For the \(m\)-th group, the frequency-domain features $\hat{F}_{T_t}$ via multi-scale low-pass filter (msLF) can be expressed as:
\begin{equation}
\small
\label{eq:mlSF}
\hat{F}_{T_t} = \operatorname{msLF}(F_{T_t})
= \Phi(\operatorname{AvgPool}_{k_m \times k_m}(F_{T_t})),
\end{equation}
where $\operatorname{AvgPool}_{k_m \times k_m}$ denotes the average pooling function with a kernel size of $k_m \times k_m$, and $\Phi(\cdot)$ represents the bilinear interpolation operation. 

For the student feature \(F_S\), we design a learnable low-pass filter, which consists of a multi-scale low-pass filter, a convolutional downsampling module, and a depthwise separable convolution (DSConv). The learnable student feature $\hat{F}_{S}$ in frequency domain is expressed as:
\begin{equation}
   \hat{F}_{S} = \operatorname{Conv}_{3 \times 3}(\operatorname{Concat}[\operatorname{DownSample}_{s \times s}(F_{S}), \operatorname{msLF}(F_{S})]), 
\end{equation}
where \(\text{DownSample}_{s \times s}\), Concat and \(\text{Conv}_{3 \times 3}\) indicates the convolutional downsampling module, feature concatenation operation, and \(3 \times 3\) convolution operation, respectively.
After obtaining the frequency-domain teacher expert and student features with msLF transformation, the total loss of SFA using the losses in Eq.~\eqref{total_loss} is rewritten as: $\mathcal{L}_\text{SFA}=\mathcal{L}_\text{MMD} + \mathcal{L}_\text{MSE}$.

\textbf{Compact Feature Alignment.}
CFA projects the feature set of the fourth layer (the penultimate layer before the fully connected layer) of all models into a compact high-dimensional spherical space \(\mathcal{Z}\).
In this space, the student model can learn degrading-related hierarchical knowledge from different pre-trained experts through spatial-domain feature alignment.
Considering the heterogeneity among models, the output feature dimensions of their encoders may differ.
Before performing CFA, a \(1 \times 1\) convolutional kernel is appended to the end of each encoder to adjust the output features of different encoders to the same dimension.
In the spherical space  \(\mathcal{Z}\), the total loss for spatial CFA is computed using the MMD and MSE losses as: $\mathcal{L}_\text{SFA}=\mathcal{L}_\text{MMD} + \mathcal{L}_\text{MSE}$.

\textbf{Uncertainty-aware Decoupled Distillation.}
To address the expert bias for the output prediction, we propose an uncertainty-aware distillation module that dynamically transfers both global and local knowledge from each expert to the student network. 
Given the logits maps \(L_{T_t} \in \mathbb{R}^{C\times H\times W}\) and \(L_S \in \mathbb{R}^{C\times H\times W}\) from the expert \(T_t\) and student \(S\), we apply spatial partitioning \(\mathcal{P}(w, w)\) at multiple scales \(w \in W = \{1, 2, 4, ..., w_{\text{max}}\}\).
For each partitioned cell \(Z(w,n)\) at scale \(w\) (where \(n \in N_w = \{1, 4, 16, ..., w^2\}\)), the accumulated logits are:
\begin{equation}
    \psi_{T_t}(w,n) = \frac{1}{w^2}\sum_{(j,k)\in Z(w,n)} L_{T_t}(j,k), \quad 
\psi_S(w,n) = \frac{1}{w^2}\sum_{(j,k)\in Z(w,n)} L_S(j,k).
\end{equation}

Then, we devise an uncertainty coefficient $U_{T_t}$ that incorporates the teacher's prediction confidence. 
For each scale-region pair (w,n), building upon the decoupled knowledge distillation paradigm~\cite{zhao2022decoupled}, the UDD loss can be defined as:  
\begin{equation}
\begin{aligned}
\mathcal{L}_{\text{UDD}}(w,n) = (2 + U_{T_t}) \cdot \mathcal{L}_{\text{TCKD}} + (1 - U_{T_t}) \cdot \mathcal{L}_{\text{NCKD}}. \\
\end{aligned}
\end{equation}
Here, the loss components are defined as: $\mathcal{L}_{\text{TCKD}} = \|\sigma(\psi_{T_t}(w,n)) - \sigma(\psi_S(w,n))\|^2_2 $, $\mathcal{L}_{\text{NCKD}} = \|\psi_{T_t}(w,n) - \psi_S(w,n)\|^2_2$. $\sigma$ denotes the $softmax$ function.
$U_{T_t} = 1 - \max(\sigma(\psi_{T_t}(w,n))) \in [0,1]$, which quantifies prediction ambiguity by measuring the deviation from a one-hot distribution. The \((2 + U_{T_t})\) term amplifies supervision on ambiguous regions (\(U_{T_t} \rightarrow 1\)) where expert predictions tend to be unreliable, while \((1 - U_{T_t})\) maintains precise logit alignment for task-agnostic regions (\(U_{T_t} \rightarrow 0\)).

\textbf{Training Loss Function.}
The total training loss can be described as:
\begin{equation}
\small
\mathcal{L}_\text{Total}= \underbrace{ \mathcal{L}_{\text{cls}}}_{\text{Classification}} + \underbrace{\alpha \cdot (\mathcal{L}_{\text{SFA}}+\mathcal{L}_{\text{CFA}})}_{\text{feature alignment}} +\underbrace{\beta \cdot \sum_{w \in W} \sum_{n \in N_w} \mathcal{L}_{\text{UDD}}(w,n)}_{\text{uncertainty-aware decoupled distillation}},
\end{equation}
where $\mathcal{L}_{\text{cls}}$ is the \textit{cross-entropy} loss and \(\alpha, \beta\) balance the loss components. 
\section{Experiments}
\subsection{Setups} 
\textbf{Datasets and Metrics.}
We evaluate our UMKD on two widely used datasets: SICAPv2~\cite{silva2020going} for histology prostate grading and APTOS~\cite{aptos2019-blindness-detection} for fundus image grading.
Four essential metrics are adopted for performance comparison: overall accuracy (OA), mean accuracy (mAcc), weighted F1-score (F1), and mean absolute error (MAE). The \textbf{bold} and \underline{underline} indicate the best and the second-best performance in each sub-dataset test.

\textbf{Implementation details.} We conduct all experiments on two real-world challenging tasks: \textit{sources-imbalanced} distillation and \textit{target-imbalanced} distillation.
In \textit{sources-imbalanced} distillation, the expert models are trained on imbalanced source datasets, and distillation is performed on class-balanced target datasets.
In \textit{target-imbalanced} distillation, the expert models are trained on balanced source datasets, and distillation is performed on class-imbalanced target datasets.
Specifically, we generate balanced subsets from original imbalanced datasets through random sampling, including SICAPv2-balanced (2500, 2222, 2500, 948) and APTOS-balanced (600, 370, 300, 193, 295) for each grading category.
The training, validation, and test sets for all datasets are split in an 8:1:1 ratio, and data augmentation techniques such as random cropping and flipping are employed to expand the dataset.
Notably, color jittering is excluded due to the sensitivity of pathological images to color variations, as random color injection could disrupt pathological features.
For model training, we utilize two ImageNet pre-trained ResNet50 models as expert models and one ImageNet pre-trained ResNet18 model as the student model.

\textbf{Baselines.}
We compare UMKD with a variety of the most representative SOTA methods, including feature-based (FitNet~\cite{romero2014fitnets} and RKD~\cite{park2019relational}), as well as logits-based (KD~\cite{HintonVD15}, DKD~\cite{zhao2022decoupled} and the latest SDD~\cite{li2024enhancing,wei2024scaled}).
In addition, we also report the results of ResNet~\cite{he2016deep} trained on each dataset as a benchmark.
\subsection{Comparison Results}
\subsubsection{Results on SICAPv2 Grading.}
\begin{table*}[tp!]
\caption{Prostate cancer grading using individually trained ResNet models (Top), feature-based KD models (Middle), and KD models (Bottom). 
} 
\centering
\begin{tabular}{l|cccc|cccc}
\toprule
\multirow{2}{*}{Methods} & \multicolumn{4}{c|}{Source-imbalanced KD (\%)}  & \multicolumn{4}{c}{Target-imbalanced KD (\%)} \\
\cline{2-9} &  OA$\uparrow$&  mAcc$\uparrow$ &  F1 $\uparrow$ &  MAE $\downarrow$ &  OA$\uparrow$&  mAcc$\uparrow$ &  F1 $\uparrow$ &  MAE$\downarrow$       \\ \midrule
Resnet50 ($Exp_1$) &91.53 &89.48 & 91.47 & 0.1098 
& 89.19 & 89.44 &89.13 & 0.1332 \\
Resnet50 ($Exp_2$) & 92.05 &89.88 & 91.93 & 0.1100 
&89.71 & 89.78 & 89.61 & 0.1322 \\
Resnet18 ($Stu$) & 89.58 & 89.06 & 89.49 & 0.1463 
&90.36 & 88.11 & 90.23 & 0.1318 \\\midrule
FitNet~\cite{romero2014fitnets} & 78.78 & 78.41 & 78.42 & 0.3136   & 79.06 &55.77 & 77.12 & 0.3694  \\

RKD~\cite{park2019relational} &88.54 &88.24 &88.44 &0.1690  &90.79 &88.30 &90.67 &0.1369 \\
\midrule

KD~\cite{HintonVD15} &\underline{89.06} &\underline{88.39} &\underline{88.97} &0.1624 
&\underline{90.97} &\underline{89.44} & \underline{90.90} & \underline{0.1368}  \\
DKD~\cite{zhao2022decoupled}  &86.91 & 85.01 & 86.68 &0.1739 
&89.19 &87.11 &89.12 &0.1476 \\
SDD~\cite{wei2024scaled}  & 87.82 & 86.66 & 87.67 & \underline{0.1594} 
& 89.93 &88.81 &89.85 & 0.1447  \\
\rowcolor{lm_purple_low} UMKD & \textbf{91.02} & \textbf{90.23} & \textbf{90.94} &\textbf{0.1294} 
& \textbf{91.75} & \textbf{90.72} &\textbf{91.72}& \textbf{0.1199} \\
\midrule
\rowcolor{lm_purple_low}\textbf{$\triangle$} 
& \textcolor{green!50!black}{+3.20} 
& \textcolor{green!50!black}{+3.57} 
& \textcolor{green!50!black}{+3.27} &\textcolor{green!50!black}{+0.0300} &\textcolor{green!50!black}{+1.82} &\textcolor{green!50!black}{+1.91} &\textcolor{green!50!black}{+1.87} & \textcolor{green!50!black}{+0.0248}\\

\bottomrule
\end{tabular}
\label{tab_sicapv2}
\end{table*}
Our UMKD outperforms all previous logits-based and feature-based KD baselines across all metrics, achieving a new state-of-the-art (SOTA) performance as shown in Table~\ref{tab_sicapv2}.

In the \textit{sources-imbalanced} KD task, UMKD achieves the highest overall accuracy (OA = 91.02\%), mean accuracy (mAcc = 90.23\%), and weighted F1 score (F1 = 90.94\%), while also attaining the lowest mean absolute error (MAE = 0.1294). Specifically, compared to SDD~\cite{wei2024scaled}, UMKD improves the mean accuracy by 3.57\%, with all performance gains highlighted in \textcolor{green!50!black}{green}.
Similarly, in the \textit{target-imbalanced} KD task, UMKD achieves SOTA performance, with mean accuracy exceeding 90.72\%.
This consistent superiority of UMKD across both tasks highlights its robustness and generalizability in diverse distillation settings.

Notably, the \textit{sources-imbalanced} KD is more challenging than the \textit{target-imbalanced} task due to the inherent biases in the expert models' knowledge, which are trained on imbalanced datasets.
Our UMKD explicitly quantifies and mitigates this imbalance bias, ensuring a more robust and reliable distillation process. 
In summary, our UMKD not only reduces the number of model parameters (ResNet18 vs. ResNet50) but also significantly enhances model performance, making it a promising tool for improving the accuracy and reliability of prostate cancer grading and diagnosis. 
\subsubsection{Results on APTOS Grading.}
\begin{table*}[tp!]
\caption{Fundus image grading using individually trained ResNet models (Top), feature-based KD models (Middle), and KD models (Bottom).
} 
\centering
\begin{tabular}{l|cccc|cccc}
\toprule
\multirow{2}{*}{Methods} & \multicolumn{4}{c|}{Sources-imbalanced KD (\%)}  & \multicolumn{4}{c}{Target-imbalanced KD (\%)} \\

\cline{2-9} &  OA$\uparrow$&  mAcc$\uparrow$ &  F1 $\uparrow$ &  MAE $\downarrow$ &  OA$\uparrow$&  mAcc$\uparrow$ &  F1 $\uparrow$ &  MAE$\downarrow$       \\ \midrule
Resnet50 ($Exp_1$) &82.34 &66.33 & 80.63 & 0.2478 & 72.66 & 63.74 &72.18 & 0.4001 \\
Resnet50 ($Exp_2$) & 82.81 &65.77 & 81.89 & 0.2421 &72.66 & 63.74 & 72.51 & 0.3936 \\
Resnet18 ($Sup$) & 74.21 & 67.18 & 73.94 & 0.4192 &82.34 & 70.56 &81.04 & 0.2521 \\\midrule
FitNet~\cite{romero2014fitnets} & 67.57 & 59.12 &66.57 & 0.5704   & 79.06 &55.77 &77.12 &0.3694  \\

RKD~\cite{park2019relational} & \underline{74.61} & \underline{67.15} & \underline{74.37} & 0.4203
& \textbf{85.00} & 69.75 & \textbf{84.38} & \underline{0.2482} \\
\midrule

KD~\cite{HintonVD15} &73.04 &66.07 &72.77 &0.4448 &83.43 & 71.56 & 83.50 & 0.2578  \\
DKD~\cite{zhao2022decoupled}  &70.70 &61.74 &69.83 &0.4018 
&81.87 &73.95 &82.43 &0.2743  \\
SDD~\cite{wei2024scaled}  & 73.44 &65.07 &72.62 & \underline{0.4017} 
&83.12 &73.55 &82.83 &0.2662  \\
\rowcolor{lm_purple_low} UMKD & \textbf{74.61} & \textbf{67.33} & \textbf{74.43} &\textbf{0.3589} 
& \underline{83.91} & \textbf{74.38} &\underline{84.03}& \textbf{0.2476} \\
\midrule
\rowcolor{lm_purple_low}\textbf{$\triangle$} & \textcolor{green!50!black}{+1.17} & \textcolor{green!50!black}{+2.26} & \textcolor{green!50!black}{+1.81} & \textcolor{green!50!black}{+0.0428} & \textcolor{green!50!black}{+0.79} & \textcolor{green!50!black}{+0.83} & \textcolor{green!50!black}{+1.20} & \textcolor{green!50!black}{+0.0186} \\

\bottomrule
\end{tabular}
\label{tab_aptos}
\end{table*}
We evaluate the performance of UMKD on the more challenging APTOS dataset, where data is more imbalanced and expert annotation is biased due to the inherent complexity of fundus images.

As shown in Table~\ref{tab_aptos}, UMKD achieves superior performance in both \textit{sources-imbalanced} and \textit{target-imbalanced} KD tasks compared to existing methods.
Specifically, for \textit{sources-imbalanced} KD task, our proposed UMKD consistently outperforms all baselines, achieving the highest overall accuracy (OA = 74.61\%), mean accuracy (mAcc = 67.33\%), and weighted F1 score (F1 = 74.43\%), while attaining the lowest mean absolute error (MAE = 0.3589). 
These results demonstrate UMKD's capability to quantify the prediction bias in expert models induced by imbalanced training data. By leveraging the uncertainty of the expert models' outputs, the student model can adaptively adjust the weights of knowledge transfer, thereby ensuring a more robust and reliable distillation process while significantly enhancing model performance.

In the \textit{target-imbalanced} distillation task, UMKD achieves SOTA performance compared to all logits-based methods, attaining the highest mean accuracy (mAcc = 74.38\%) and maintaining strong performance across other metrics (OA = 83.91\%, F1 = 84.03\%).
As shown in 7-th row, UMKD yields suboptimal results in OA and F1 compared to the feature-based RKD~\cite{park2019relational}, yet it significantly outperforms RKD in mACC (74.38\% vs. 69.75\%).
We attribute this to the following reasons: first, RKD optimizes the angular momentum to compute mini-batch distances between sample triplets, encouraging samples of the same class to cluster closer together.
While this approach achieves the highest overall accuracy, it tends to favor majority classes in \textit{target-imbalanced} distillation tasks due to the inherent class imbalance within batches.
In contrast, UMKD addresses this bias by explicitly measuring and mitigating the imbalance with uncertainty, achieving a better trade-off between overall accuracy and mean accuracy.
\subsection{Ablation Study}
    


\begin{table}[tp!]
    \caption{Ablation study of SFA, CFA, and UDD modules on SICAPv2 dataset.}
    
    \centering
    \begin{tabular}{ccc|cccc|cccc}
    \toprule
    \multicolumn{3}{c|}{Methods} & \multicolumn{4}{c|}{Sources-imbalanced KD (\%)}  & \multicolumn{4}{c}{Target-imbalanced KD (\%)} \\
    \cmidrule(lr){4-7} \cmidrule(lr){8-11}
    SFA & CFA & UDD & OA$\uparrow$ & mAcc$\uparrow$ & F1 $\uparrow$ & MAE $\downarrow$ & OA$\uparrow$ & mAcc$\uparrow$ & F1 $\uparrow$ & MAE$\downarrow$ \\ \midrule
    & & & 87.82 & 86.66 & 87.67 & 0.1594 & 89.93 &88.81 &89.85 & 0.1447\\
    &  \checkmark& \checkmark &90.69   &89.92   &90.58   &0.1314   &91.62    &90.54   &91.58   &0.1258  \\
    \checkmark & &\checkmark &90.36   &89.68   &90.25   &0.1355   &91.44    &90.93   &91.40   &0.1275  \\
    \checkmark & \checkmark &  & 88.15 & 86.84 &  87.97 & 0.1566  &  90.06 & 89.15 &  90.03 & 0.1416\\
    \rowcolor{lm_purple_low}\checkmark &\checkmark & \checkmark &\textbf{91.02} & \textbf{90.23} & \textbf{90.94} &\textbf{0.1294} & \textbf{91.75} & \textbf{90.72} &\textbf{91.72}& \textbf{0.1199} \\
    \bottomrule
    \end{tabular}
    \label{tab:ablation_study}
\end{table}
We conduct an ablation study to analyze the contributions of the three components of UMKD.
The results on the SICAPv2 dataset, as shown in Table~\ref{tab:ablation_study}, demonstrate that all components are essential for achieving high performance in both \textit{source-imbalanced} and \textit{target-imbalanced} grading tasks.
The absence of the SFA and CFA components leads to significant performance degradation in UMKD. This is because the task-agnostic structural features and task-specific semantic features are not effectively decoupled. In disease image grading, pathological features are often localized within specific image regions. Relying solely on global features can obscure these critical classification-related features, reducing the model's ability to accurately grade diseases.

Similarly, without UDD, the student model cannot dynamically adjust the knowledge transfer weights, leading to the worst distillation performance.
This highlights the importance of uncertainty-aware mechanisms in mitigating bias propagation and ensuring robust knowledge transfer.
It is worth stating that our ablation experiments on the APTOS dataset yield consistent conclusions, but are not reported due to space limitations.

\section{Conclusion}
We proposed a UMKD framework to address the challenge of data imbalance in grading tasks.
By integrating the frequency-domain SFA and spatial-domain CFA modules, UMKD effectively decoupled task-agnostic structural features from task-specific semantic features. 
The UDD mechanism further enhanced robustness by dynamically adjusting knowledge transfer weights based on expert knowledge uncertainties and mitigated biases induced by imbalanced data and model heterogeneity.
Extensive experiments on fundus and prostate cancer datasets have demonstrated that UMKD achieved state-of-the-art performance in both source-imbalanced and target-imbalanced scenarios.
These results highlighted UMKD's ability to ensure reliable and fair knowledge transfer, making it a practical and scalable solution for real-world disease image grading.
Future work will focus on extending UMKD to other medical imaging tasks and improving its interpretability for clinical adoption.
\bibliographystyle{splncs04}
\bibliography{miccai2025}

\begin{thebibliography}{10}
\providecommand{\url}[1]{\texttt{#1}}
\providecommand{\urlprefix}{URL }
\providecommand{\doi}[1]{https://doi.org/#1}

\bibitem{bulten2022artificial}
Bulten, W., Kartasalo, K., Chen, P.H.C., Str{\"o}m, P., Pinckaers, H., Nagpal, K., Cai, Y., Steiner, D.F., Van~Boven, H., Vink, R., et~al.: Artificial intelligence for diagnosis and gleason grading of prostate cancer: the panda challenge. Nature medicine  \textbf{28}(1),  154--163 (2022)

\bibitem{cheng2023robust}
Cheng, Y., Ying, H., Hu, R., Wang, J., Zheng, W., Zhang, X., Chen, D., Wu, J.: Robust image ordinal regression with controllable image generation. arXiv preprint arXiv:2305.04213  (2023)

\bibitem{claudio2024mapping}
Claudio~Quiros, A., Coudray, N., Yeaton, A., Yang, X., Liu, B., Le, H., Chiriboga, L., Karimkhan, A., Narula, N., Moore, D.A., et~al.: Mapping the landscape of histomorphological cancer phenotypes using self-supervised learning on unannotated pathology slides. Nature Communications  \textbf{15}(1), ~4596 (2024)

\bibitem{dai2021deep}
Dai, L., Wu, L., Li, H., Cai, C., Wu, Q., Kong, H., Liu, R., Wang, X., Hou, X., Liu, Y., et~al.: A deep learning system for detecting diabetic retinopathy across the disease spectrum. Nature communications  \textbf{12}(1), ~3242 (2021)

\bibitem{gao2024collaborative}
Gao, S., Fu, Y., Liu, K., Gao, W., Xu, H., Wu, J., Han, Y.: Collaborative knowledge amalgamation: Preserving discriminability and transferability in unsupervised learning. Information Sciences  \textbf{669},  120564 (2024)

\bibitem{gao2023contrastive}
Gao, S., Fu, Y., Liu, K., Han, Y.: Contrastive knowledge amalgamation for unsupervised image classification. In: International Conference on Artificial Neural Networks. pp. 192--204. Springer (2023)

\bibitem{gao2024ka}
Gao, S., Fu, Y., Liu, K., Xu, H., Wu, J.: Ka$^{2}$ er: Knowledge adaptive amalgamation of experts for medical images segmentation. arXiv preprint arXiv:2410.21085  (2024)

\bibitem{gao2023bayeseg}
Gao, S., Zhou, H., Gao, Y., Zhuang, X.: Bayeseg: Bayesian modeling for medical image segmentation with interpretable generalizability. Medical Image Analysis  \textbf{89},  102889 (2023)

\bibitem{hao2024one}
Hao, Z., Guo, J., Han, K., Tang, Y., Hu, H., Wang, Y., Xu, C.: One-for-all: Bridge the gap between heterogeneous architectures in knowledge distillation. Advances in Neural Information Processing Systems  \textbf{36} (2024)

\bibitem{he2016deep}
He, K., Zhang, X., Ren, S., Sun, J.: Deep residual learning for image recognition. In: Proceedings of the IEEE conference on computer vision and pattern recognition. pp. 770--778 (2016)

\bibitem{HintonVD15}
Hinton, G., Vinyals, O., Dean, J.: Distilling the knowledge in a neural network. arXiv preprint arXiv:1503.02531  (2015)

\bibitem{aptos2019-blindness-detection}
Karthik, Maggie, Dane, S.: Aptos 2019 blindness detection. \url{https://kaggle.com/competitions/aptos2019-blindness-detection} (2019), kaggle

\bibitem{li2024enhancing}
Li, L., Li, X.C., Ye, H.J., Zhan, D.C.: Enhancing class-imbalanced learning with pre-trained guidance through class-conditional knowledge distillation. In: Forty-first International Conference on Machine Learning (2024)

\bibitem{litjens2017survey}
Litjens, G., Kooi, T., Bejnordi, B.E., Setio, A.A.A., Ciompi, F., Ghafoorian, M., Van Der~Laak, J.A., Van~Ginneken, B., S{\'a}nchez, C.I.: A survey on deep learning in medical image analysis. Medical image analysis  \textbf{42},  60--88 (2017)

\bibitem{mohan2023drfl}
Mohan, N.J., Murugan, R., Goel, T., Roy, P.: Drfl: federated learning in diabetic retinopathy grading using fundus images. IEEE Transactions on Parallel and Distributed Systems  \textbf{34}(6),  1789--1801 (2023)

\bibitem{park2019relational}
Park, W., Kim, D., Lu, Y., Cho, M.: Relational knowledge distillation. In: Proceedings of the IEEE/CVF conference on computer vision and pattern recognition. pp. 3967--3976 (2019)

\bibitem{porwal2020idrid}
Porwal, P., Pachade, S., Kokare, M., Deshmukh, G., Son, J., Bae, W., Liu, L., Wang, J., Liu, X., Gao, L., et~al.: Idrid: Diabetic retinopathy--segmentation and grading challenge. Medical image analysis  \textbf{59},  101561 (2020)

\bibitem{romero2014fitnets}
Romero, A., Ballas, N., Kahou, S.E., Chassang, A., Gatta, C., Bengio, Y.: Fitnets: Hints for thin deep nets. arXiv preprint arXiv:1412.6550  (2014)

\bibitem{silva2020going}
Silva-Rodr{\'\i}guez, J., Colomer, A., Sales, M.A., Molina, R., Naranjo, V.: Going deeper through the gleason scoring scale: An automatic end-to-end system for histology prostate grading and cribriform pattern detection. Computer methods and programs in biomedicine  \textbf{195},  105637 (2020)

\bibitem{wang2023ord2seq}
Wang, J., Cheng, Y., Chen, J., Chen, T., Chen, D., Wu, J.: Ord2seq: Regarding ordinal regression as label sequence prediction. In: Proceedings of the IEEE/CVF International Conference on Computer Vision. pp. 5865--5875 (2023)

\bibitem{wei2024scaled}
Wei, S., Luo, C., Luo, Y.: Scaled decoupled distillation. In: Proceedings of the IEEE/CVF Conference on Computer Vision and Pattern Recognition. pp. 15975--15983 (2024)

\bibitem{xie2021survey}
Xie, X., Niu, J., Liu, X., Chen, Z., Tang, S., Yu, S.: A survey on incorporating domain knowledge into deep learning for medical image analysis. Medical Image Analysis  \textbf{69},  101985 (2021)

\bibitem{zhao2022decoupled}
Zhao, B., Cui, Q., Song, R., Qiu, Y., Liang, J.: Decoupled knowledge distillation. In: Proceedings of the IEEE/CVF Conference on computer vision and pattern recognition. pp. 11953--11962 (2022)

\end{thebibliography}
\end{document}